\documentclass[a4paper]{article}
\usepackage{amsmath,amstext,amsgen,amsbsy,amsopn,amsfonts,amssymb}
\usepackage{easybmat}
\usepackage{graphics}
\usepackage[pdftex]{graphicx}
\usepackage{epsfig}
\usepackage{hyperref}
\usepackage{amsthm}
\usepackage{float}
\usepackage{bm}
\usepackage{color}
\usepackage{algorithmic}
\usepackage{algorithm}
\usepackage{quoting}
\usepackage{flushend}
\usepackage{balance}
\usepackage{multicol}
\usepackage[resetlabels]{multibib}

\begin{document}
\title{\textbf{Incorporating Deep Features in the Analysis of Tissue Microarray
Images}}

\author{
Donghui Yan$^{\dag}$, Timothy W. Randolph$^{\ddag}$, Jian Zou$^{\$}$, Peng Gong$^{\P}$
\vspace{0.1in}\\
$^\dag$Mathematics, University of Massachusetts Dartmouth, MA\vspace{0.05in}\\
$^\ddag$Fred Hutchinson Cancer Research Center, WA\vspace{0.05in}\\
$^\$$Mathematical Sciences, Worcester Polytechnic Institute, MA\vspace{0.05in}\\
$^\P$Department of ESPM, University of California Berkeley, CA \\[0.02in]Department of Earth System Science, Tsinghua University, China\\[0.05in]
}

\date{\today}
\maketitle


\begin{abstract}
\noindent
Tissue microarray (TMA) images have been used increasingly often in cancer studies and
the validation of biomarkers. TACOMA---a cutting-edge automatic scoring algorithm for TMA 
images---is comparable to pathologists in terms of accuracy and repeatability. Here we
consider how this algorithm may be further improved. Inspired by the recent success of
deep learning, we propose to incorporate representations learnable through computation.
We explore representations of a group nature through unsupervised learning, e.g.,
hierarchical clustering and recursive space partition. Information carried by clustering or 
spatial partitioning may be more concrete than the labels when the data are heterogeneous, 
or could help when the labels are noisy. The use of such information could be viewed as 
regularization in model fitting. It is motivated by major challenges in TMA image
scoring---heterogeneity and label noise, and the {\it cluster} assumption in
semi-supervised learning. Using this information on TMA images of breast cancer, we
have reduced the error rate of TACOMA by about $6\%$.
Further simulations on synthetic data provide insights on when such representations
would likely help. Although we focus on TMAs, learnable representations of this type
are expected to be applicable in other settings.
\end{abstract}

\section{Introduction}
\label{section:introduction}
\begin{figure}[htp]
\centering
\begin{center}
\hspace{0cm}
\includegraphics*[scale=0.36,clip]{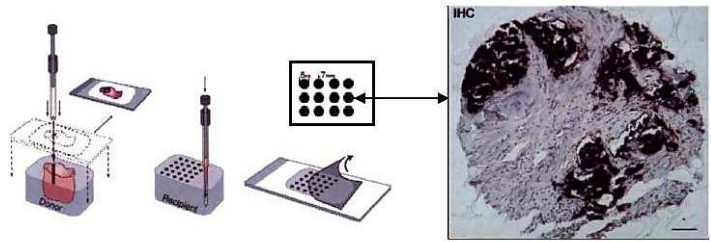}
\end{center}
\abovecaptionskip -2pt
\caption{{\it An illustration of the TMA technology (the left half of the image was taken from \cite{TACOMA}). 
Small tissue cores are first extracted 
from tumor blocks, and stored in archives which are frozen or preserved with formalin. Then thin 
slices of tissues are sectioned from the tissue core. A tissue slide is formed by an array of hundreds 
of tissue sections (possibly from different patients). Biomarkers are  then applied to the tissue 
sections (which then typically show darker colors). A TMA image is then captured for each tissue 
section from a high-resolution microscope. }} 
\label{figure:tmaTech}
\end{figure}

The tissue microarray (TMA) technology was developed during the last three decades \cite{CampNR2008,Kononen1998,WanFF1987} 
as a high-throughput technology for the evaluation of histology-based laboratory tests. A particularly
desirable feature of TMAs is that they allow the immunohistochemical (IHC) 
staining of hundreds of sections all at once, thus standardizing many variables
involved. A TMA slide is an array of hundreds of thin tissue sections cut from
small-core biopsies (less than $1$ mm in diameter). Such biopsies are taken 
from cell lines, or archives of frozen or formalin-fixed paraffin-embedded tissues.
These arrayed sections are then stained and mounted on a TMA slide, which will be
viewed with a high-resolution microscope. A TMA image is produced from each tissue section. 
Figure~\ref{figure:tmaTech} is an illustration of the TMA technology. 
\\
\\
A standard approach to quantify the qualitative IHC readings is for a pathologist to provide a single-number score to each spot
which summarizes the pattern of staining as it relates to specific types of cells.  For example, a protein marker that is highly 
expressed in cancerous cells will exhibit a qualitatively different pattern than a marker that is less indicative of cancer and may 
exhibit non-specific staining. These scores serve as a convenient proxy to study the tissue images, given the complexity (staining 
patterns are not localized in positions, shape or size) and potential high dimensionality of tissue images (a TMA image typically 
has a size of $1504 \times 1440$ pixels). They have been used for a wide array of applications, including the validation 
of biomarkers, assessment of therapeutic targets, analysis of clinical outcome \cite{Hassan2008}, tumor progression analysis 
\cite{BeckSL2011},
and the study of genomics and proteomics (``imaging genetics") \cite{HibarKS2011}. 
The use of TMAs in cancer biology has increased dramatically in recent years \cite{CampNR2008,Giltnane2004,Hassan2008,Voduc2008}.
Particularly, since TMAs allow the rapid assessment of DNA, RNA and protein expression on large tissue samples, they are 
emerging as the standard tool for the validation of diagnostic and prognostic biomarkers \cite{Hassan2008}.
\\
\\
The inherent variability and subjectivity with manual scoring \cite{Bentzen2008,Berger2005,CampNR2008,Chung2002, 
DivitoCamp2005,Giltnane2004, Hsu2002, Thomson2001, Vrolijk2003, Walker2006} 
of TMA images, as well as the demand for reproducible large-scale high-throughput call for the automatic 
scoring of TMA images. A number of commercial tools have been developed, including ACIS (ChromaVision 
Medical Systems), Ariol (Applied Imaging), TMAx (Beecher Instruments) and TMALab II (Aperio) for IHC, 
and the AQUA method \cite{CampRimm2002} (HistRx, Inc.) for fluorescent labeled images. However, most 
are difficult to tune and the resulting models are sensitive to many variables such as IHC staining quality,
background antibody binding, hematoxylin counterstaining, and the color and hue of chromogenic reaction 
products used to detect antibody binding. 
\\
\\
This work extends TACOMA---an automatic scoring algorithm for tissue images that is robust against various 
factors such as variability in the image intensity and staining patterns etc \cite{TACOMA}. While TACOMA 
achieves a scoring accuracy comparable to a trained pathologist on a number of tumor and biomarker 
combinations \cite{TACOMA}, naturally, one would wonder if it is possible to make further progress. One 
source of inspiration comes from the recent advance in deep learning \cite{HintonSalakhutdinov2006, LeCunBengioHinton2015}, 
especially in the area of image classification \cite{KrizhevskySutskeverHinton2012, SimonyanZisserman2015, SzegedyLJ2015}.
For TMA images, however, the huge training set required by deep neural networks, typically at the magnitude 
of millions, is hard to obtain in reality (techniques such as transfer learning \cite{TorreyShavlik2009} may help, but 
still it is not easy to get large enough training sample). In a typical TMA database, for example, the Stanford 
TMAD database \cite{Marinelli2007}, the size of the training set associated with any particular biomarker is merely 
in the order of hundreds. 
\\
\\
There are several factors that would limit the availability of TMA images. While natural images---the type of images 
that deep learning has had huge success on---or their labels can be easily obtained by web scraping or crowd-sourcing, 
it is much harder for TMA images which have to be acquired from human body and captured by high-resolution 
microscopes and high-end imaging devices. Moreover, the labelling of TMA images is typically done by pathologists. 
In terms of classification, the natural and TMA images are of a completely different nature. A natural image typically 
consists of a hierarchy of well-defined image objects,
which form important high-level features for image categorization. In contrast, the scoring of TMA images is not 
about how the staining pattern looks like, rather the ``severity and spread" of the pattern matters, i.e., it 
concerns some global property and requires considerable expertise. The sample size is further limited 
by the fact that TMA images are scored by biomarkers or cancer types; there are over 100 cancer types 
according to the US National Cancer Institute \cite{nciCancer}. 
\\
\\
What lesson can we learn from deep learning? Rather than a tool for building a powerful classifier with deep layers 
of neural networks, we view the essence of deep learning as a way of finding a suitable representation (possibly 
hierarchical) for the underlying problem through computation. Such a representation would otherwise be hidden 
from manual feature engineering. In particular, we are able to use unsupervised learning to find features of a group nature, which along with existing 
features used by TACOMA, leads to improved performance in scoring. This was motivated by known major challenges
in the scoring of TMA images---heterogeneity and label noise, and inspired by the {\it cluster} assumption in semi-supervised 
learning \cite{ChapelleWeston2003, Zhu2008}. As such new features are typically beyond usual feature engineering 
and had to be found by computation, we term those deep features; of course by ``deep" also means we had inspirations 
from deep learning and the new features are produced from existing features. For this reason and due to the intimate 
connection of our approach to TACOMA, we term our approach {\it deepTacoma}. 
\\
\\
The organization of the remainder of this paper is as follows. We describe the TACOMA algorithm in Section~\ref{section:tacoma}. 
In Section~\ref{section:method} we discuss our method and some new classes of feature representations. In Section~\ref{section:experiments}, we present our experiments and results. We conclude with Section~\ref{section:conclusion}.
\section{The TACOMA algorithm}
\label{section:tacoma}
In this section, we will briefly describe the TACOMA algorithm. This will provide a basis to understand 
{\it deepTacoma} which extends upon TACOMA. To ensure consistency in notations, we begin with an 
introduction of notations following \cite{YanBickelGong2017,TACOMA}. Note that the scoring systems 
\cite{Marinelli2007} adopted in practice typically use a small number of discrete values, such as $\{0,1,2,3\}$, 
as the {\it score} (or {\it label}) for TMA images. We formulate the scoring of TMA images as a classification 
problem, following \cite{TACOMA}. 
\\
\\
The primary challenge in TMA image analysis is the lack of easily-quantified criteria for scoring: features 
of interest are not localized in position, shape or size. There are no ``landmarks" and no hope of ``image 
registration" for comparing features. Rather, this problem is truly a challenge about quantifying {\it qualitative} 
properties of the TMA images. The key insight that underlies TACOMA is that in spite of heterogeneity, 
TMA images exhibit strong statistical regularity in the form of visually observable textures or {\it staining 
patterns}. In TACOMA, such patterns are captured by an important image statistics---the gray level co-occurrence 
matrix (GLCM).
\subsection{The gray level co-occurrence matrix}
\begin{figure}[ht]
\centering
\begin{center}
\hspace{0cm}
\includegraphics*[scale=0.34,clip]{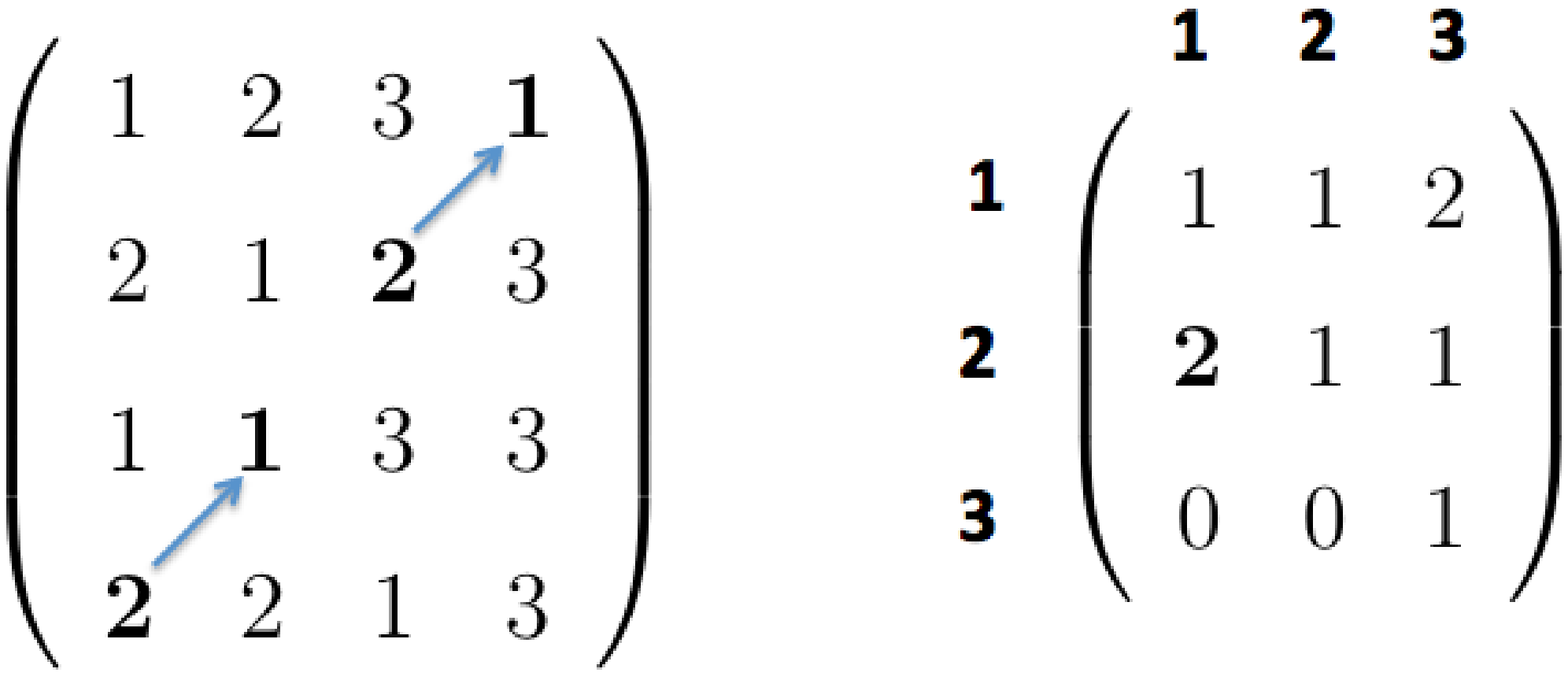}\\\vspace{-0.1in}{\bf (a)}\\
\includegraphics*[scale=0.41,clip]{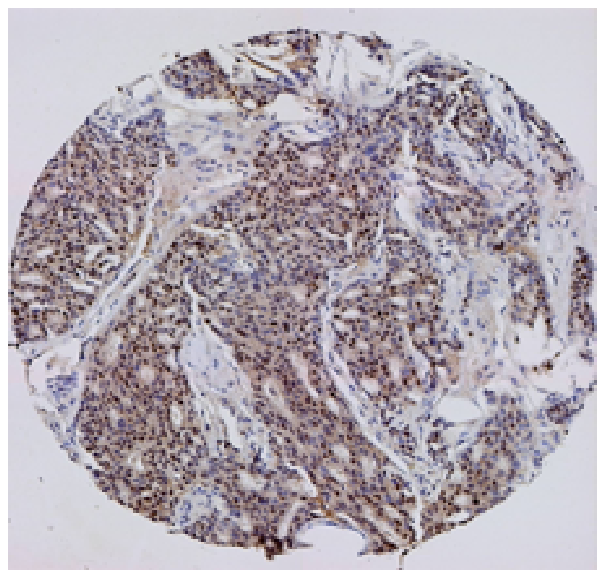}
\includegraphics*[scale=0.41,clip]{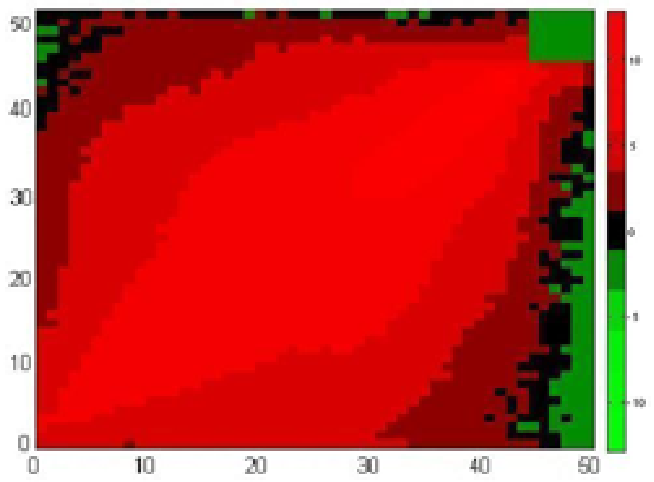}\\
{\bf (b)}
\end{center}
\caption{{\bf Example images and GLCMs}. ~{\bf (a)} A toy image and its GLCM.  {\it The toy image 
is a $4 \times 4$ image with a $3 \times 3$ GLCM for $\sim=(\nearrow,1)$.} ~{\bf (b)} A TMA image 
(left panel) and the heatmap of its GLCM (right panel, in log scale, and taken from \cite{TACOMA}). 
{\it In the right panel, the axis labels (0-50) indicate the normalized pixel values in a TMA 
image; the color (scale indicated by a color bar) of the heatmap represents 
the value of entries in the GLCM. } } \label{figure:glcms}
\end{figure}

The GLCM of an image is a matrix of counting statistics about the spatial 
pattern of neighboring pixels. It can be crudely viewed as a ``histogram" according to 
a certain spatial relationship. It was proposed by Haralick \cite{haralick1979} and
has been proven successful in a variety of applications
\cite{GongMarceau1992,haralick1979, Lloyd2004,TACOMA}. 
The GLCM is defined with respect to a particular spatial relationship described below. 
\\
\\
\textbf{Definition \cite{YanBickelGong2017,TACOMA}.} The spatial relationship between 
a pair of pixels in image $I$ involves their relative position and spatial distance. The 
set $\Re$ of spatial relationships of interest is defined as
\begingroup
\setlength\abovedisplayskip{6pt}
\setlength{\belowdisplayskip}{6pt}
\begin{equation*}
\Re = \mathit{D} \otimes \mathit{L} 
=  \{\nearrow,~\searrow,~ \nwarrow, ~\swarrow, ~\downarrow,
~\uparrow, ~\rightarrow, ~\leftarrow\} \otimes \{1,...,d\}
\end{equation*}
\endgroup
where $D$ is the set of possible directions, and $L$ is the distance
between the pair of pixels along the direction. 
\\
\\
\textbf{Definition \cite{YanBickelGong2017,TACOMA}.} For a given spatial relationship $\sim \in \Re$, the GLCM 
for an image (or a patch) is defined as (assume the number of gray levels in the image is $N_g$)
\begin{quoting}
\quotingsetup{vskip=-1pt}
\noindent A $N_g \times N_g$ matrix such that its $(a,b)$-entry
counts the number of times two pixels, $P_1 \sim P_2$, and their gray values are $a
~\mbox{and}~ b$, respectively, for $a, b \in \{1,2,...,N_g\}$.
\end{quoting}
Note that, an image can have multiple GLCMs, with each corresponding to a particular spatial
relationship. The definition of GLCM is illustrated in Figure~\ref{figure:glcms} with a toy and a real TMA image 
(taken from \cite{TACOMA}). For a good balance of computational efficiency and discriminative power, we take 
$N_g=51$ and apply uniform quantization \cite{GrayNeuhoff1998} over the $256$ gray levels in our application.

\subsection{An algorithmic description of TACOMA}
The TACOMA algorithm is particularly simple to describe. First, all TMA images are converted to 
their GLCM representations. Then the training set (GLCMs and their respective scores) is fed to 
some training algorithm to obtain a trained classifier. The trained classifier will be applied to get 
scores for TMA images in the test set. Random Forests (RF) \cite{RF} is chosen as the training 
algorithm due to its exceptional performance in many classification tasks that involve high dimensional 
data \cite{caruanaKY2008}. 
\\
\\
Denote the training sample by $(I_1,Y_1),...,(I_n,Y_n)$ where $I_i$'s are images and $Y_i$'s are 
scores (thus $Y_i \in \{0,1,2,3\}$). Let $I_{n+1}, ..., I_{n+m}$ be new TMA images that one wish to 
score (i.e., the test set has a size of $m$). Additionally, let $Z_1,...,Z_{l}$ denote the small set of 
`representative' image patches; $l=5$ in TACOMA \cite{TACOMA}. TACOMA is described as 
Algorithm~\ref{algorithm:tacoma}.
\begin{algorithm}
\caption{The TACOMA algorithm}
\label{algorithm:tacoma}
\begin{algorithmic}[1]
\STATE For each image patch $Z_i$, compute its GLCM matrix
$Z_i^g,i=1,...,l$; 
\STATE Build feature mask $\mathcal{M}_i$ as the set of indices for which entries of $Z_i^g$ surpass 
a threshold $\tau_i$, and set $\mathcal{M}=\cup_{i=1}^l \mathcal{M}_i$; 
\FOR {$i=1$ \TO $n+m$} 
\STATE Compute GLCM of image $I_i$ and keep only entries in the
index set $\mathcal{M}$; 
\STATE Denote the resulting matrix by $X_i$; \ENDFOR 
\STATE Feed $\cup_{i=1}^n \{(X_i,Y_i)\}$ to RF to obtain a classification rule $\hat{f}$;
\STATE Apply $\hat{f}$ to $X_i$ to obtain score for image $I_i, i=n+1,...,n+m$.
\end{algorithmic}
\end{algorithm}
\noindent Here $\tau_i$ is a threshold value used to filter unimportant 
features. In particular, we set it to be the median of all entries in GLCM matrix $Z_i^g$ for $i=1,...,l$. 
Steps 1-2 are optional; these are used to create a {\it mask} through which all GLCMs are filtered. The 
mask is created 
from a small number TMA image patches selected by pathologists that would reflect important aspects 
when they manually score the images. This is where the pathologists could incorporate their domain 
expertise in the TACOMA algorithm. In this work, however, we will not include masking due to the lack 
of pathologists for the verification of representative image patches; it is included in the description mainly 
to be consistent with how TACOMA was described in \cite{TACOMA}.

\section{The {\it deepTacoma} method}
\label{section:method}
The main idea of our method is to look for some ``good" representation of the TMA images for the purpose of scoring. 
Here ``good" means such a representation can lead to information beyond the straightforward use of GLCM features. 
This is essentially a feature engineering problem (see, for example, \cite{GuyonElisseeff2003}), and there are many 
possibilities one could explore. Our strategy is to look for those representations of a group nature which reflects how 
close data points are to each other. This is motivated by practical success of the {\it cluster} assumption in semi-supervised 
learning \cite{ChapelleWeston2003, Zhu2008}, as well as known major challenges in developing scoring algorithms for 
TMA images---heterogeneity and label noise. Alternatively, one may consider using regularized mixture models proposed 
recently by Gao and his coauthors \cite{GaoShenOmbao2018} to deal with those challenges, but that is beyond the scope 
of the present work. Our approach is really a problem-driven approach---we directly target at those specific known challenges 
and seek representations that are informative towards them. Our approach is implemented as two classes of features, one 
generated from clustering (including K-means and hierarchical clustering) and the other based on recursive space partition. 
\\
\\
The representations we explore are a group property that relates different data points and is beyond what may be revealed 
by features of individual data points alone. Therefore we expect such representations would lead to additional 
information that may help in the scoring of TMA images. One may argue that the label (or score) along with the TMA 
images would have already captured such information. We note, however, that TMA images receiving the same score 
could still be highly heterogeneous. Thus, information carried by clustering or space partition may be more concrete 
than that by the labels. Heterogeneity is itself, in certain sense, a group property. Including those features related 
to grouping may help in directing the algorithm to build sub-models to deal with heterogeneous data as appropriate. 
Moreover, TMA labels are typically noisy. Different pathologists may score differently, and the same pathologist 
may give different scores to the same image at different scoring sessions \cite{TACOMA}. The information carried by 
clustering or space partition would likely help against label noise, in a similar way as the {\it cluster} assumption in 
semi-supervised learning would do to compensate the scarcity of labeled instances: many data instances 
do not have a label, and information from those labeled instances could be borrowed through the group 
property. One could view the information revealed by clustering or space partition as a type of regularization in model 
fitting. Thus either a more stable model, or a model with better accuracy, would be expected. 
\begin{figure}[h]
\centering
\begin{center}
\hspace{0cm}
\includegraphics[scale=0.32,clip]{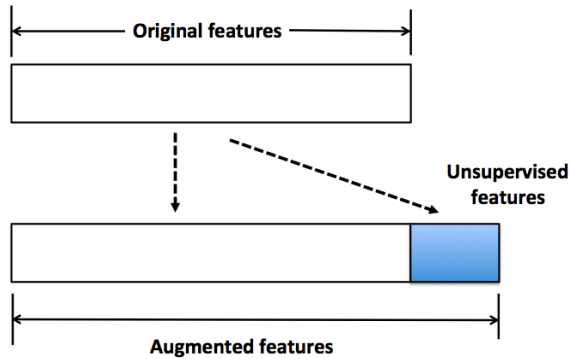}
\end{center}
\caption{\it Illustration of deep features. The unsupervised features (or deep features) are generated from 
the original features, which together form the augmented features to be used for classification tasks.} 
\label{figure:dpFeatures}
\end{figure}
\\
\\
The deep features, together with the original features, form the augmented features. This will be the input 
to a classification algorithm (RF in our case). Figure~\ref{figure:dpFeatures} is an illustration of how this is 
done. For clustering-based deep features, we use the cluster IDs as the deep features. To obtain a more 
informative set of deep features, in the case of hierarchical clustering, we generate the dendrogram first 
and then cut it at many different heights. Each height will lead to a different clustering of the data. The IDs 
obtained from clustering at all different heights form the set of deep features. For recursive space partition, 
we implement by an ensemble of random projection trees (rpTrees) \cite{RPTree} (i.e., random 
projection forests \cite{rpForests2018c}). Many rpTrees are generated, and the leaf node IDs 
from each tree form a deep feature. Instead of tree leaf node IDs, we also attempt to encode 
the path from the root to each leaf node, but that is less effective. 
\\
\\
For the rest of this section, we will briefly describe K-means clustering, hierarchical clustering, and rpTrees. 
\subsection{K-means clustering}
$K$-means clustering was developed by S. Lloyd in 1957 (but published later in 1982) \cite{lloyd1982}, and 
remains one of the simplest yet most popular clustering algorithms. 
The goal of $K$-means clustering is to split data into $K$ partitions (clusters) and assign each point to the 
``nearest" cluster (mean). A cluster mean is the center of mass of all points in a cluster, or the arithmetic 
mean of all points in a cluster; it is also called {\it cluster centroid or prototype}. The algorithm is very simple. 
Starting with a set of randomly selected cluster centers, the algorithm alternates between two steps: assign 
all the points to its nearest cluster centers, and recalculate the new cluster centers, and stops when no further 
changes are observed on the cluster centers. 
\\
\\
For a more detailed description of K-means clustering, please 
refer to the appendix (c.f., Section~\ref{section:AppendixKmeans}) or \cite{hartiganWong1979, lloyd1982}. 
\subsection{Hierarchical clustering}
Hierarchical clustering refers to a class of clustering algorithms that first organize the data in a hierarchy (called 
{\it dendrogram}), and then form clusters by cutting through the dendrogram at a certain height. Depends on 
how the hierarchy is formed, bottom up or top down, there are two types of hierarchical clustering approaches, 
agglomerative or divisive clustering. In the following, we will briefly describe them. 
\\
\\
Agglomerative clustering is a bottom-up approach. It starts by treating each data point as a singleton cluster  
(i.e., a cluster that contains only one data point). Two points (or clusters) that are the most similar are fused 
to form a bigger cluster. Then points or clusters are continually fused one-by-one in order of highest similarity 
or cluster to which they are most similar. Eventually, all points are merged to form a single ``giant" cluster. This 
produces a dendrogram to be cut through at a certain height to form clusters. 
\\
\\
Divisive clustering takes a top-down approach. Initially, all data points belong to the same 
cluster. Then, recursively, clusters are divided until each cluster contains only one data point. At each 
stage, the cluster with largest diameter (defined as the largest dissimilarity between any two points in a cluster) 
is selected for further division. To divide the selected cluster, one looks for its most disparate observations (the 
point with largest average dissimilarity to others), which initiates the so-called ``splinter group", then re-assign data points 
closer to the``splinter group" as one group and the rest as another group. The selected cluster is split into two 
smaller new clusters. 
\\
\\
For more details 
about hierarchical clustering, please refer to \cite{HTF2001, KaufmanRousseeuw1990, Ward1963}. 
\begin{figure}[ht]
\centering
\begin{center}
\hspace{0cm}
\includegraphics[scale=0.25,clip]{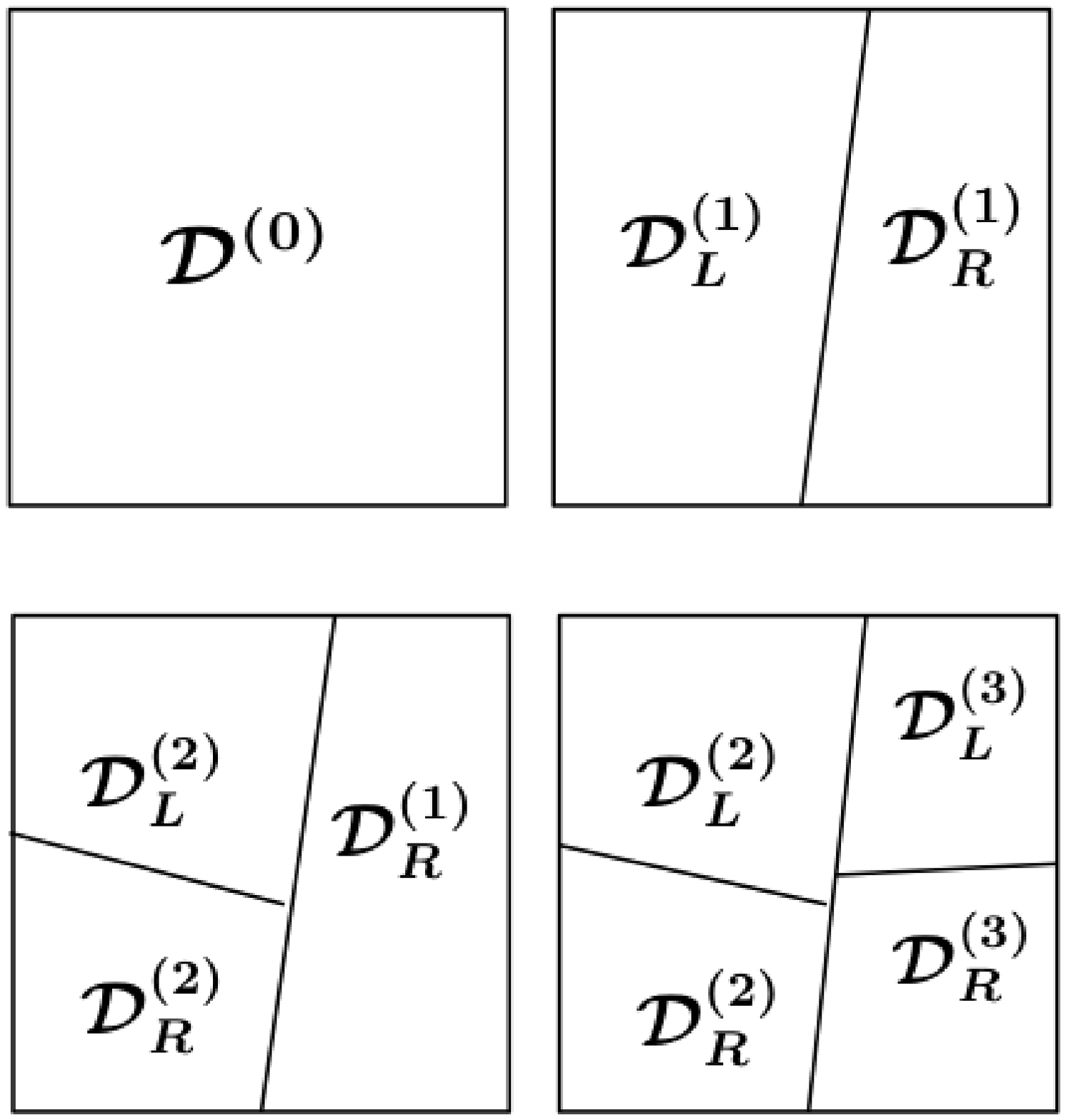}~~~~
\includegraphics[scale=0.25,clip]{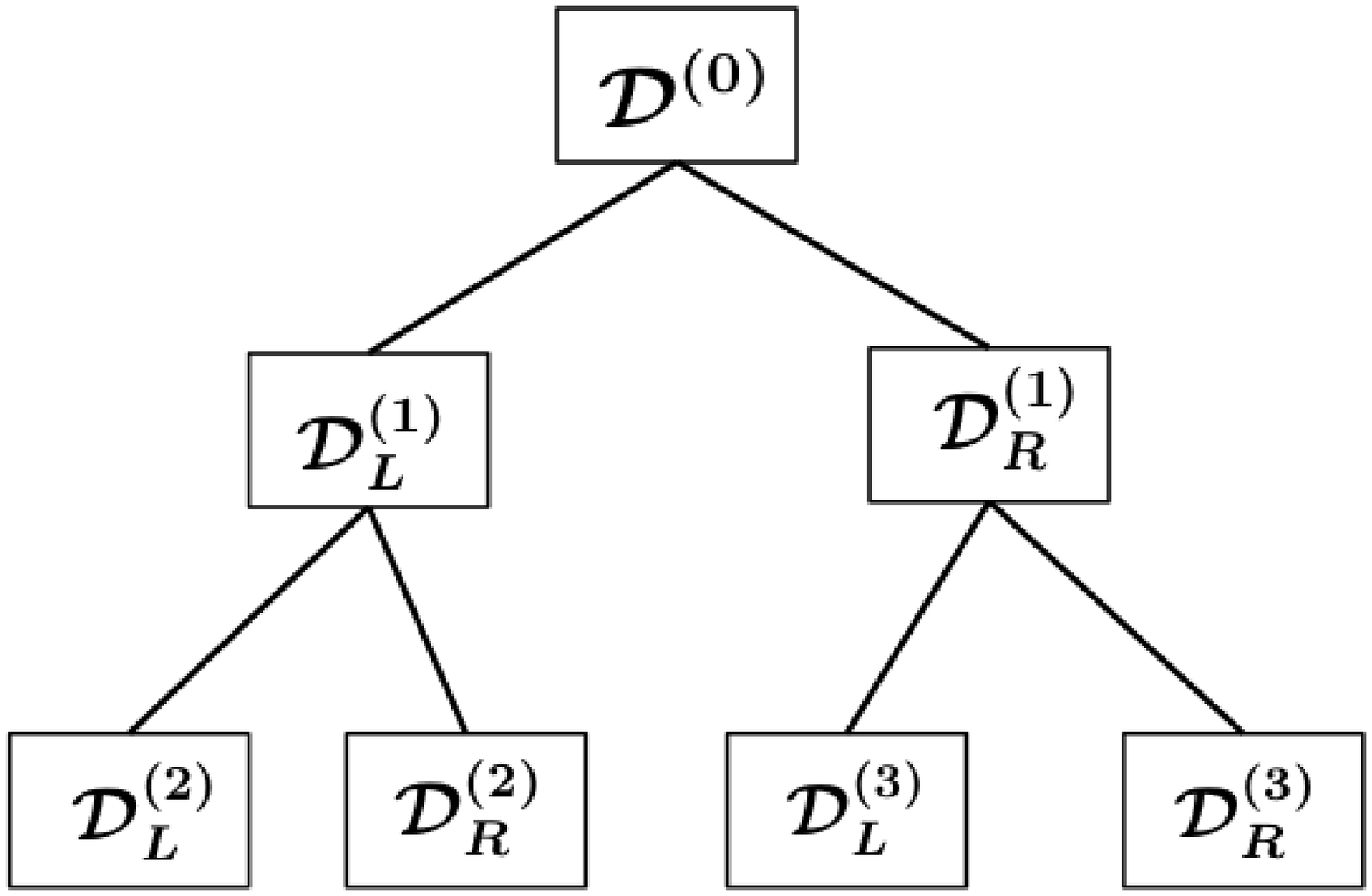}
\end{center}
\caption{\it Illustration of space partition and random projection trees. The superscripts indicate the order 
of tree node split. One starts with the root node,
$\mathcal{D}^{(0)}$, which corresponds to all the data. After the
first split, $\mathcal{D}^{(0)}$ is partitioned into its two child nodes, 
$\{\mathcal{D}_{L}^{(1)}, \mathcal{D}_{R}^{(1)}\}$.
The second split partitions the left child node,
$\mathcal{D}_{L}^{(1)}$, into its two child nodes,
$\{\mathcal{D}_{L}^{(2)}, \mathcal{D}_{R}^{(2)}\}$. The third split would be on
node $\mathcal{D}_{R}^{(1)}$ which leads to
two new child nodes, $\{\mathcal{D}_{L}^{(3)}, \mathcal{D}_{R}^{(3)}\}$. This process
continues until a stopping criterion is met.} 
\label{figure:rpTree}
\end{figure}
\subsection{Recursive space partition}
Another type of representation is based on recursive space partition. It is a class of methods widely used 
in many applications \cite{FriedmanBentleyFinkel1977, HuntMarkStoll2006, YanDavis2018,YanHuangJordan2009} 
that organizes the data points according to their proximity in a recursive fashion. 
The recursive space partition is typically implemented with a popular data structure, the k-d tree \cite{kdTree}. We 
use its randomized version, rpTrees \cite{RPTree}. One advantage of rpTrees over kd-tree is its ability to adapt to 
the geometry of the underlying data and readily overcomes the curse of dimensionality, according to \cite{RPTree}. 
The rpTrees starts with the entire data set, $\mathcal{D}^{(0)}$, as the root node. The split on the root node results 
in two child nodes, on each of which the same splitting procedure applies recursively until some stopping criterion 
is met. For example, the node becomes too small (i.e., contains two few data points). Data points falling into the same
leaf node will be `similar' to each other. An illustration of recursive space partition via a tree is shown in Figure~\ref{figure:rpTree}. 
\\
\\
The algorithmic implementation of rpTrees uses a queue, $\mathcal{W}$, of working nodes to implement rpTrees. 
Initially the queue contains only the root node, $D$, which corresponds to the given data set. In each iteration, a 
node is picked from the queue, then split (or no processing if it is smaller than a predefined size, $n_s$), and the 
resulting child nodes are pushed into the queue. This process continues until there are no more working nodes in 
the queue. Let $P_{\vec{r}}(a)$ denote the projection of point $a$ onto line $\vec{r}$. The algorithm will return $t$, 
the rpTrees to be built from $D$. The pseudo code for the algorithm is described as Algorithm~\ref{algorithm:neighGen}.
\begin{algorithm}
\caption{\it~~rpTree(D)}
\label{algorithm:neighGen}
\begin{algorithmic}[1]
\STATE Let $D$ be the root node of tree $t$; 
\STATE Initialize the set of working nodes $\mathcal{W} \leftarrow \{D\}$; 
\WHILE {$\mathcal{W}$ is not empty}
	\STATE Randomly pick $W \in \mathcal{W}$ and set $\mathcal{W} \leftarrow \mathcal{W} - \{W\}$; 
	\IF{$|W| < n_s$} 
		\STATE Skip to the next round of the while loop; 
	\ENDIF 
    	\STATE Generate a random line $\vec{r}$;  
	\STATE Project points in $W$ onto $\vec{r}$, let $M=\{P_{\vec{r}}(w): w \in W \}$; 
	\STATE Let $a=\min(M)$ and $b=\max(M)$; 
	\STATE Generate a splitting point $s$ uniformly over interval $[a,b]$; 
	\STATE Split node $W$ by $W_L=\{a: P_{\vec{r}}(a) < s\}$ and $W_R=\{a: P_{\vec{r}}(a) \geq s\}$; 
	\STATE $W.left \leftarrow W_L$ and $W.right \leftarrow W_R$; 
	\STATE Update the working set by $\mathcal{W} \leftarrow \mathcal{W} \cup \{W_L, W_R\}$; 
\ENDWHILE
\STATE return(t); 
\end{algorithmic}
\end{algorithm} 
\section{Experiments}
\label{section:experiments}
We conduct experiments on both synthetic data and TMA images. The TMA images are the data that our 
methods are primarily targeting at. The synthetic data are generated from Gaussian mixtures, and serve 
the purpose of gaining insights on when our `deep' features may help. Our approach is motivated 
by the intuition that features leveraging the group property may be useful when the labels are noisy or when 
the data are heterogeneous, and we have expressly created simulation scenarios (i.e., Gaussian mixtures 
$\mathcal{G}_1$ and $\mathcal{G}_2$ in Section~\ref{section:expGMixture}) for these. 
\\
\\
For simplicity, we consider 0-1 loss for classification throughout, and use the error rate on the test sample 
as our performance metric. For K-means clustering, we use the R package {\it kmeans}, and for hierarchical 
clustering, we use three different R packages, including {\it hclust}, {\it diana}, and {\it agnes}. For rpTrees, 
an ensemble of rpTrees (i.e., random projection forests) are generated and each tree corresponds to a new deep 
feature; we use an R implementation for random projection forests \cite{rpForests2018c}. The ensemble size 
is related to the model complexity of the resulting class of classifiers; we can increase the ensemble size when 
the training sample size increases. RF is chosen as the classifier. According 
to \cite{TACOMA}, RF is far superior to support vector machines (SVM) \cite{CortesVapnik1995} and boosting 
methods \cite{AdaBoost2} in scoring TMA images. This is likely due to the high data dimension (2601 when 
using GLCM) and potential label noise in the data. RF has strong built-in capability in feature selection and 
noise-resistance, while SVM and boosting methods are typically prune to those. This is also supported by 
related work on the segmentation of tissue images \cite{HolmesKapelner2009}, and many large scale simulation 
studies \cite{caruanaKY2008}. In all our experiments, we fix the number of trees in RF to be 100 (adequate 
by our experience, and no attempt is made in finding the best number), and the number
of tries in selecting variables for node split is chosen from $\{[\sqrt{p}],[2\sqrt{p}]\}$ where $p$ is the number of 
features in the data. Also for all simulations, a randomly selected half of the data are used for training and the 
rest for test, and results are averaged over $100$ runs. 
\\
\\
For the rest of this section, we present 
details about our experiments on the Gaussian data and the TMA images.
\subsection{Gaussian mixtures}
\label{section:expGMixture}
Three different Gaussian mixtures, $\mathcal{G}_1, \mathcal{G}_2$ and $\mathcal{G}_3$, are considered. 
$\mathcal{G}_1$ and $\mathcal{G}_2$ correspond to the usual Gaussian mixture data, and heterogeneous 
data, respectively, and $\mathcal{G}_3$ uses the covariance matrix estimated from the GLCM matrix of TMA
images used in our experiment. Gaussian mixture $\mathcal{G}_1$ is specified as
\begin{eqnarray}
\label{eq:gm} \frac{1}{2}\mathcal{N}(\bm{\mu},\Sigma) + \frac{1}{2}
\mathcal{N}(\bm{-\mu},\Sigma),
\end{eqnarray}
where `1/2's indicate that half of the data are generated from $\mathcal{N}(\bm{\mu},\Sigma)$ and half
from $\mathcal{N}(\bm{-\mu},\Sigma)$. Here $\mathcal{N}(\bm{\mu},\Sigma)$
stands for Gaussian distribution with mean $\bm{\mu} \in \mathbb{R}^p$ and covariance matrix $\Sigma$. 
We take $p=40$, and $\bm{\mu}=(0.3,...,0.3)^T$. The covariance matrix $\Sigma$ is defined such that its 
$(i,j)$-entry is given by 
\begin{equation*}
\Sigma_{ij}=\rho^{|i-j|}, ~\mbox{for $\rho \in \{0.1,0.3,0.5\}$.}
\end{equation*}
If the data is from $\mathcal{N}(\bm{\mu},\Sigma)$, then we assign it a class label `1' otherwise a label `2'. 
The sample size for all of $\mathcal{G}_1, \mathcal{G}_2$ and $\mathcal{G}_3$ are 1000. To see the effect 
of label noise, we randomly select a proportion of $\epsilon$ of the training instances and 
flip their labels, i.e., change from `1' to `2' or from `2' to `1'.
\\
\\
Our naming convention and experimental settings are as follows. `RF' indicates results by RF on original 
data; `hClustering' for results by RF on original data augmented by features derived from hierarchical clustering; 
`rpTrees' for results by RF on original data augmented by features derived from an ensemble of rpTrees; `K-means' for results 
by RF on original data augmented by features derived from K-means clustering. Such a convention is followed 
through our experiments. For K-means clustering, the best results are reported when the number of clusters 
varies from $\{30, 40,...,120\}$. For hierarchical clustering, the number of clusters ranges from $[10,60]$ and 
all three different hierarchical clustering procedures are used. Note that here clustering is used as a tool to
extract latent structures from the data by grouping similar or nearby data points, the exact number of clusters 
is no longer as important as in the usual clustering setting. The main goal is to ensure that the grouping is fine 
``enough", and meanwhile each group has a sufficient number of data points. The same applies to recursive 
space partitions. For rpTrees, we try different ensemble size in the set $\{200, 400, 600, 800\}$, and find the 
difference small with 800 doing slightly better; no attempt is made in obtaining the best results. The maximum 
size of a node is fixed at $20$. The experimental results are summarized in Table~\ref{table:accuracydpGaussianNoise}. 
\begin{table}[htb]
\begin{center}
\begin{tabular}{c|c|c|c|c|c}
    \hline
\bf{$\rho$}         & \bf{$\epsilon$}                        &\bf{RF}  &\bf{K-means} &\bf{hClustering} &\bf{rpTrees} \\
\hline
0.1                     &0              	&$8.18\%$    &$7.68\%$     &$5.16\%$    &$5.82\%$        \\
                          &0.1              	&$9.25\%$    &$8.90\%$     &$5.52\%$    &$6.32\%$        \\
                     &0.2              	&$11.16\%$   &$10.71\%$  &$6.91\%$    &$8.06\%$       \\
                     &0.3              	&$15.28\%$   &$15.04\%$  &$11.21\%$  &$12.25\%$     \\
\hline
0.3                  &0              	        &$11.55\%$   &$11.08\%$  &$9.26\%$  &$9.51\%$         \\       
                     &0.1              	&$12.32\%$   &$12.16\%$  &$9.68\%$  &$9.98\%$        \\
                     &0.2              	&$13.77\%$   &$13.53\%$  &$11.15\%$  &$11.61\%$     \\
                     &0.3              	&$18.09\%$   &$17.69\%$  &$16.17\%$  &$15.58\%$    \\
\hline
0.5                     &0              	&$15.81\%$    &$15.73\%$  &$14.47\%$  &$14.38\%$   \\
                     &0.1              	&$16.73\%$   &$16.44\%$  &$15.43\%$  &$14.97\%$    \\
                     &0.2              	&$17.83\%$   &$17.56\%$  &$17.09\%$  &$16.43\%$    \\
                     &0.3              	&$22.17\%$   &$21.87\%$  &$21.98\%$  &$19.88\%$    \\
\hline
\end{tabular}
\end{center}
\caption{\it{Error rates on Gaussian mixture $\mathcal{G}_1$. 
}} \label{table:accuracydpGaussianNoise}
\end{table}
\\
\\
It can be seen from Table~\ref{table:accuracydpGaussianNoise} that, in all cases, both the hierarchical clustering and
the rpTrees based approaches lead to reduced error rates while the gain by K-means clustering is marginal (indicating 
that more refined structural information may be required). Moreover, when the label noise is moderate, for example when 
$\epsilon=0.1$, the reduction in error rate is often more significant than other cases (including the case without label noise,
i.e., $\epsilon=0$). 
When $\rho$ is small, that is, individual features in the data are less correlated, deep features tend to lead to more 
substantial improvement in classification performance. This is probably because, in such settings, one can get higher 
quality unsupervised features (as a result of better clustering or space partitions). 
\\
\\
Gaussian mixtures $\mathcal{G}_2$ is specified as 
\begin{eqnarray*}
\label{eq:gm2} \frac{1}{4}\mathcal{N}(\bm{\mu_1},\Sigma) + \frac{1}{4}
\mathcal{N}(\bm{\mu_2},\Sigma) + \frac{1}{4}\mathcal{N}(\bm{-\mu_1},\Sigma) + \frac{1}{4}
\mathcal{N}(\bm{-\mu_2},\Sigma),
\end{eqnarray*}
which indicates that a quarter of the data are generated from each of the 4 Gaussians. Same as $\mathcal{G}_1$, 
we take $p=40$. The Gaussian mixture centers are fixed as $\bm{\mu_1}=(0.5,...,0.5,0,...,0)^T$ and 
$\bm{\mu_2}=(0,...,0,0.5,...,0.5)^T$, where for both $\bm{\mu_1}$ and $\bm{\mu_2}$, exactly half of the components 
are $0$. The covariance matrices are the same as for $\mathcal{G}_1$. If the data is generated from either 
$\mathcal{N}(\bm{\mu_1},\Sigma)$ or $\mathcal{N}(\bm{\mu_2},\Sigma)$, then we assign it a class label `1' otherwise 
a label `2'. This produces heterogeneous data in the sense that data with the same class label may be from different 
Gaussians. The results are reported in Table~\ref{table:accuracydpGaussianHetero} with similar patterns as in 
Table~\ref{table:accuracydpGaussianNoise}. 
\begin{table}[htb]
\begin{center}
\begin{tabular}{c|c|c|c|c|c}
    \hline
\bf{$\rho$}         & \bf{$\epsilon$}                        &\bf{RF}  &\bf{K-means} &\bf{hClustering} &\bf{rpTrees}    \\
\hline
0.1                     &0              	&$12.69\%$   &$12.45\%$  &$9.89\%$  &$10.36\%$           \\
                          &0.1              	&$13.64\%$   &$13.55\%$  &$10.50\%$  &$11.53\%$         \\
                     &0.2              	&$15.63\%$    &$15.42\%$ &$12.38\%$  &$13.40\%$        \\
                     &0.3              	&$20.53\%$   &$20.18\%$  &$17.37\%$  &$18.48\%$         \\
\hline
0.3                  &0              	         &$15.69\%$   &$15.91\%$ &$14.11\%$  &$14.14\%$         \\
                     &0.1              	&$17.28\%$   &$16.79\%$ &$14.95\%$  &$15.22\%$         \\
                     &0.2              	&$18.76\%$   &$18.61\%$  &$16.67\%$  &$16.95\%$         \\
                     &0.3              	&$23.41\%$   &$23.03\%$&$22.39\%$  &$21.37\%$         \\
\hline
0.5                     &0              	&$19.56\%$   &$20.49\%$ &$19.85\%$  &$18.07\%$         \\
                     &0.1              	&$20.65\%$   &$21.33\%$ &$20.50\%$  &$19.14\%$         \\
                     &0.2              	&$22.63\%$    &$23.02\%$ &$23.07\%$  &$21.08\%$        \\
                     &0.3              	&$26.35\%$   &$26.67\%$ &$26.67\%$  &$24.44\%$         \\
\hline
\end{tabular}
\end{center}
\caption{\it{Error rates on Gaussian mixtures $\mathcal{G}_2$. 
}} \label{table:accuracydpGaussianHetero}
\end{table}
\\
\\
Gaussian mixture $\mathcal{G}_3 \in \mathbb{R}^{2601}$ is specified as $\frac{1}{2}\mathcal{N}(-\bm{\mu},\Sigma)+\frac{1}{2}\mathcal{N}(\bm{\mu},\Sigma)$, 
where $\Sigma$ is estimated from the GLCM of all TMA images used in our experiment. 
Table~\ref{table:accuracydpG3} shows the error rate by RF and that with additional 
features generated by K-means clustering, hierarchical clustering, and rpTrees, respectively. 
\begin{table}[htb]
\begin{center}
\begin{tabular}{c|r|r|r|r}
    \hline
$\epsilon$                        &\textbf{RF}~~~  &~\textbf{K-means}~ &~\textbf{hClustering}~ &~\textbf{rpTrees}~ \\
\hline
0.1              	&$1.58\%$~    &$1.48\%$~     &$1.18\%$~    &$1.10\%$~        \\
0.2              	&$3.42\%$~   &$3.24\%$~      &$3.06\%$~    &$2.40\%$~       \\
0.3              	&$9.48\%$~   &9.12$\%$~      &$8.24\%$~    &$7.68\%$~       \\
0.4				&$26.50\%$~  &$25.70\%$~   &$26.16\%$~  &$25.94\%$~     \\
\hline
\end{tabular}
\end{center}
\caption{\it{Error rates on Gaussian mixture $\mathcal{G}_3$. 
}} \label{table:accuracydpG3}
\end{table}
While deep features by K-means clustering barely improve the results, those by rpTrees 
yield notable improvement following a similar pattern as that for $\mathcal{G}_1$ and $\mathcal{G}_2$ 
(i.e., results improved when the label noise is moderate). Here, K-means or hierarchical clustering probably 
suffer from the high dimensionality of the data to which rpTrees is more resistant, a desirable property of
rpTrees \cite{RPTree}. 
\\
\\
Note here all Gaussian mixtures $\mathcal{G}_{1-3}$ use a common covariance matrix for their 
mixture components. In statistics and machine learning, it is not unusual to assume a common 
covariance matrix for Gaussian mixtures. For example, the liner discriminant analysis (LDA) arises 
from such an assumption. According to Hastie, Tibshirani and Friedman \cite{HTF2001}, 
in terms of decision boundary, the difference between LDA and quadratic 
discriminant analysis (QDA) is small, and both perform well on an amazingly large and diverse set 
of classification tasks. In the STATLOG project \cite{MichieSTC1995}, LDA was among the top three 
classifiers for 7 of the 22 datasets, QDA among the top three for four datasets. Indeed many published 
work assume a common covariance matrix for Gaussian mixtures; see, for example, \cite{BickelLevina2004,Dasgupta1999, FanFan2008,CF}.

\subsection{Applications on TMA images}
\label{section:experimentsTMA}
The TMA images are taken from the Stanford Tissue Microarray Database, or STMAD (see \cite{Marinelli2007} 
and \texttt{http://tma.stanford.edu/}). TMAs corresponding to the biomarker, estrogen receptor (ER), for breast 
cancer tissue are used since ER is a known well-studied biomarker. Each image is assigned a score (i.e., label) from $\{0,1,2,3\}$. 
The scoring criteria are:  `$0$' indicating a definite negative (no staining of tumor cells), `$3$' a definitive positive 
(most cancer cells show dark nucleus staining), `$2$' for positive (a small portion of tumor cells show staining or 
a majority show weak staining), and `$1$' indicates ambiguous weak staining in a small portion of tumor cells, or 
unacceptable image quality. 
\\
\\
There are totally $695$ TMA images for ER in the Stanford database. 
The GLCM for $(\nearrow,3)$ is used. Different choices of direction 
and distance of interaction for spatial relationship were explored in \cite{TACOMA}, and $(\nearrow,3)$ shows 
the greatest discriminating power when ER as a biomarker is used for breast cancer. The pathological interpretation 
is that, the distance of interaction is related to the size of the staining pattern for the biomarker and 
cancer type, and the staining pattern is approximately rotationally invariant (thus the choice of direction is not 
as important). Indeed when more spatial relationships are included or combined, the changes in the results 
are negligible. The deep features, either by clustering or rpTrees, are obtained for all the images. Then we fit 
{\it deepTacoma} on the training set (over the set of augmented features) and apply the fitted classifier to the 
test set. 
\\
\\
We conduct {\it three sets of experiments} on TMA images, including those on {\it deepTacoma}, when combining 
deep features generated by hierarchical clustering and rpTrees, and deep learning with TMA images. These are 
described in the next three subsections, respectively. 

\subsubsection{Experiments with {\it deepTacoma}}
\label{section:expDeepTacoma}
\begin{figure}[ht]
\centering
\begin{center}
\hspace{0cm}
\includegraphics[scale=0.5,clip]{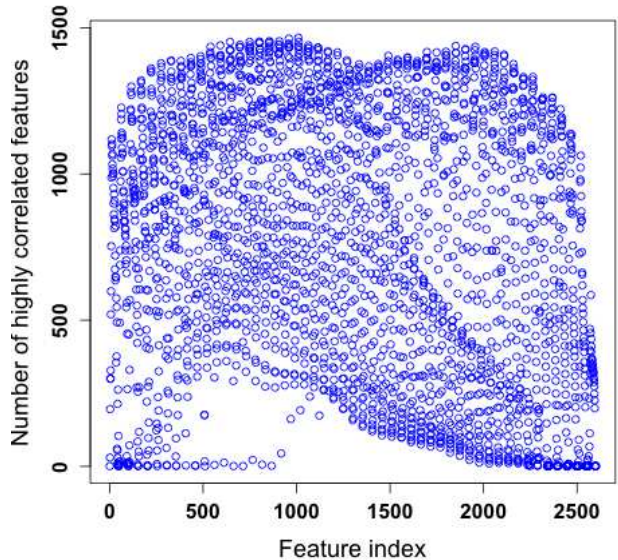}
\end{center}
\abovecaptionskip=-2pt
\caption{\it Number of highly correlated features for each of the 2601 image features. } 
\label{figure:corrTMA}
\end{figure}

The results on {\it deepTacoma} are reported in Table~\ref{table:accuracyTMA}. In the case of hierarchical clustering, 
the dendrogram is cut such that the number of groups run through $[10,40]$. Similar as the Gaussian mixture data, 
the ensemble size for rpTrees is explored from $\{200,400,600,800\}$ and a value of $600$ yields similar but slightly 
better results. An error rate at $24.79\%$ is obtained by RF on the original GLCM features (i.e., 
without using deep features). The best results are achieved when combining different hierarchical clustering algorithms 
over a range of different number of clusters, or the ensemble of rpTrees. There is about a $6\%$ reduction in error 
rates for TMA images of breast cancer, which we consider a notable improvement given that 
TACOMA algorithm already achieves a performance at the level of a trained pathologist and that progress in
this field is typically incremental in nature. 
\begin{table}[htb]
\begin{center}
\begin{tabular}{c|c|c}
    \hline
\bf{Deep features}         &\bf{\# clusters}               &\bf{Error rate}\\
                                      &\bf{ or leaf nodes} &\\
\hline
---   &---                       &$24.79\%$\\
\hline
K-means &40                     &24.02\%\\
\hline
Diana                     &[10,40]                &$24.20\%$\\
Agenes                   &[10,40]                  &$24.14\%$\\
hclust                       &[10,40]                 &$24.29\%$\\
Agenes + Diana       &[10,40]                   &$23.77\%$\\
Agenes + hclust       &[10,40]                   &$23.71\%$\\
hclust + Diana             &[10,40]                &$23.52\%$\\
Agenes + Diana + hclust     &[10,40]          &$\bm{23.46\%}$\\
\hline
rpTrees                       &30				&$\bm{23.28\%}$\\
\hline
\end{tabular}
\end{center}
\caption{\it{Error rate in scoring TMA images. 
Note that the first row corresponds to results obtained by
RF on the original set of features (i.e., without deep features). 
}} \label{table:accuracyTMA}
\end{table}
\\
\\
One possible reason that we are not able to further improve the performance of {\it deepTacoma} is probably due 
to the fact that the image features are highly correlated. According to our simulation on synthetic data 
(c.f. Table~\ref{table:accuracydpGaussianNoise} and Table~\ref{table:accuracydpGaussianHetero}), it becomes 
challenging to use deep features to further improve the performance when the correlation is high. Figure~\ref{figure:corrTMA} 
confirms this by showing the number of ``highly correlated" features for each of the 2601 features, and for most 
of the features, such a number would be larger than 500. Here by ``highly correlated" we mean the correlation 
coefficient has its absolute value larger than 0.6. Such a high correlation among features motivates us to carry 
out a principal component analysis (PCA) \cite{Hotelling1933} 
of the TMA image data and then apply RF over the leading principal 
components. Simulations are conducted using from 2 to 100 principal components (for each the results are averaged 
over 100 runs), which explains up to 99.99\% of the total variation in the data. The lowest error 
rate was 29.28\%, achieved at around 50 principal components. This may serve as a further indication on the 
hardness of scoring TMA images (an algorithm has to detect the hidden nonlinear structures in the data 
formed by TMA images to score well). 
\subsubsection{Combining deep features}
\label{section:expDeepCombine}
Given that we have formed deep features by hierarchical clustering and by rpTrees, it is possible to 
combine these two. We explore two alternatives, leaving many other possibilities to future work.
In the first option, all deep features are added to the existing pool of GLCM features and then train the classifier. 
This leads to an error rate of 23.40\%, in between what we get by using deep features separately to train
a classifier. This is likely due to the relatively small training sample size as compared to the complexity 
of the function class for the classifiers when combining features. 
\begin{figure}[h]
\centering
\begin{center}
\hspace{0cm}
\includegraphics[scale=0.50,clip]{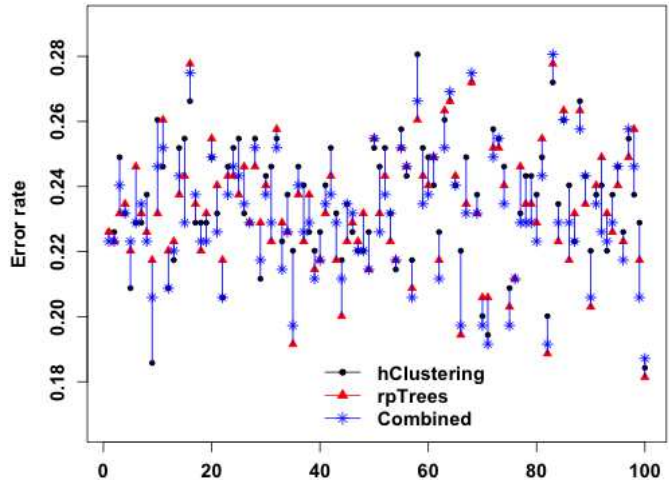}
\end{center}
\abovecaptionskip=-9pt
\caption{\it Error rates of RF with deep features by hierarchical clustering, rpTrees, and the combination of 
the two resulting RF classifiers over 100 runs. For better visualization, the three error 
rates of the same run (i.e., common training and test set) are connected vertically. } 
\label{figure:errStack}
\end{figure}
\\
\\
In the second option, we train RF classifiers with deep features by hierarchical clustering and by rpTrees 
{\it separately}, and then combine the two resulting classifiers. For a given test instance, each of the two 
classifiers gives a vote in the form of a vector of weights towards 4 classes $\{0,1,2,3\}$; denote the 
voting vectors by $v_1$ and $v_2$, respectively. The two voting vectors are combined by a simple 
linear combination $v_1 + \beta v_2$; the value of $\beta$ could be determined by cross validation. 
The label of the test instance is given by the majority class using the combined votes. As an 
example, say, $v_1=(0.38, 0.14, 0.11, 0.37), v_2=(0.28, 0.08, 0.15, 0.49)$ and $\beta=1.1$, 
the combined votes would be $(0.69,0.23,0.27,0.91)$. Individually the two classifiers would report 
a label `0' and `3', respectively, and the combined votes would report a label of `3'. Reflecting our 
belief that the classifier with deep features by rpTrees is slightly stronger, we set $\beta=1.1$. This 
leads to an error rate of $23.16\%$, marginally improving over $23.28\%$. The gain is small, however, 
if we watch over individual runs the result is actually fairly encouraging---the combined classifier has 
an error rate either close to or smaller than the best of the two in most runs. Figure~\ref{figure:errStack} 
is a scatter plot of the test set error rates by each of the two classifiers, and their combination over 100 runs.
\subsubsection{Experiments with deep learning}
\label{section:expDeepLearning}
Given the popularity of deep learning, we also carry out simulations on TMA images with deep neural networks. 
For an overview of deep neural network, please refer to \cite{GoodfellowBengioC2016}. The {\it deepnet} package
is used. The original TMA images have a size of $1504 \times 1440$, and this immediately causes problems in running 
deep neural networks due to insufficient memory of the computer (the input layer has the same number of nodes as 
the image size). We reduce the images to a number of smaller sizes, including $16 \times 16, 32 \times 32, 64 \times 64$, 
$128 \times 128$ and $256 \times 256$ (popular image datasets such as the imageNet \cite{imageNet} uses image 
size of $256 \times 256$ and MNIST \cite{MNIST} uses $28 \times 28$). The number of layers in the deep networks 
we explore range from 4 to 7 (including the input and the output layer); different number of nodes for each layers are 
explored. Table~\ref{table:accuracyDeepLearning} lists the best results obtained under different node size configurations 
that we explore for the deep neural network. For comparison, we also include results obtained by RF (on the image itself, 
just as the deep neural network does) under different image sizes. 
\begin{table}[htb]
\begin{center}
\begin{tabular}{c|c|c}
    \hline
~\textbf{Image size}~                       & ~\textbf{Deep neural network}~                        &~\textbf{RF}~   \\
\hline 
$16 \times 16$                     &$34.92\%$              	&$32.84\%$         \\
$32 \times 32$                     &$36.49\%$              	&$29.56\%$          \\
$64 \times 64$                     &$35.20\%$              	&$28.25\%$          \\
~$128 \times 128$~                 &$35.89\%$              	&~$28.82\%$~       \\
~$256 \times 256$~                 &$36.71\%$              	&~$29.70\%$~       \\
\hline
\end{tabular}
\end{center}
\caption{\it{Error rate on TMA images of different sizes by deep learning and RF. 
}} \label{table:accuracyDeepLearning}
\end{table}
It can be seen that error rates achieved by deep neural networks are higher than those by RF (both
higher than those achieved by {\it deepTacoma}). We attribute this to the small training sample size---the 
size of the training sample does not match the complexity of the function class for the deep neural 
network. 
\section{Conclusions}
\label{section:conclusion}
We propose to incorporate deep features in the analysis of TMA images. Such deep features can be learned in a small sample setting, 
which is typical of TMA images or other biomedical applications. We explore the learning of deep representations of a group nature, inspired
by the success of the {\it cluster} assumption in semi-supervised learning and known challenges in TMA images scoring---heterogeneity 
and label noise. In particular, we attempt two classes of such features, clustering-based and rpTrees-based. 
In both cases, our experiments show that incorporating such deep features lead to a further reduction of error rate by over $6\%$ 
on TACOMA for TMA images related to breast cancer. We consider this a notable improvement given that TACOMA already 
rivals trained pathologists in the scoring TMA images and the incremental nature of progress in this area. 
\\
\\
Our simulations on the Gaussian mixtures provide insights on when such deep features may help. In general, we expect that deep 
features as we propose would help when there is label noise or when the data are heterogeneous.
Note that the type of representations we have explored are of a group nature. It may be worthwhile to explore deep representations 
related to the geometry or topology of the underlying data, such as those revealed by manifold learning \cite{Cayton2008,HuoNiSmith2007} or 
topological data analysis \cite{Bubenik2015,Carlsson2009}.  

\section{Acknowledgements}
The authors are grateful to the editor, the associate editor, and the reviewers for constructive comments and suggestions.
\section{Appendix}
In this section, we will provide more details on the algorithmic implementation of K-means clustering.
\subsection{An algorithmic description of K-means clustering}
\label{section:AppendixKmeans}
Formally, given $n$ data 
points, $K$-means clustering seeks to find a partition of $K$ sets $S_1, S_2, ..., S_K$ such that the 
within-cluster sum of squares, $SS_W$, is minimized
\begin{equation}
\label{equation:kmeansFormu}
\underset{S_1,S_2,...,S_K} {\operatorname{arg\,min}} \sum_{i=1}^{K} \sum_{\mathbf x \in S_i} \left\| \mathbf x - \boldsymbol\mu_i \right\|^2,
\end{equation}
where $\mu_i$ is the centroid of $S_i, i=1,2,...,K$.
\\
\\
Directly solving the problem formulated as in \eqref{equation:kmeansFormu} is hard, as it is an integer 
programming problem. Indeed it is NP-hard \cite{ArthurVassilvitskii2006}. 
The K-means clustering algorithm is often referred to a popular implementation sketched as
Algorithm~\ref{alg:kmeans} below. For more details, one can refer to \cite{hartiganWong1979, lloyd1982}.
\begin{algorithm}[h]
\caption{\textbf{$K$-means clustering algorithm}}
\label{alg:kmeans}
\begin{algorithmic}[1]
\STATE Generate an initial set of $K$ centroids $m_1, m_2, ..., m_K$;
\STATE Alternate between the following two steps
\STATE \hspace{\algorithmicindent} Assign each point $x$ to the ``closest" cluster 
\setlength\abovedisplayskip{2pt}
\setlength\belowdisplayskip{0pt}
	\begin{equation*}     
        \arg \min_{j \in \{1,2,...,K\}} \big \| x-m_j \big \|^2;
        \end{equation*}
\STATE \hspace{\algorithmicindent} Calculate the new cluster centroids
	\begin{equation*}
	m_j^{new} = \frac{1}{\| S_j\|} \sum_{x \in S_j} x, ~~j=1,2,...,K;
        \end{equation*}
\STATE Stop when cluster assignment no longer changes.
\end{algorithmic}
\end{algorithm}
\bibliographystyle{plain}
\flushend

\end{document}